\documentclass[runningheads]{llncs}
\usepackage[T1]{fontenc}
\usepackage{graphicx}
\usepackage{comment}
\usepackage{float}
\usepackage{circledsteps}
\usepackage{hyperref}
\usepackage{listings}
\usepackage{xcolor} % for custom colors
\usepackage{listings}
\usepackage{xcolor}
\usepackage{amsmath}
\usepackage{multirow}
\usepackage{graphicx}%
\usepackage{hyperref}
\usepackage[flushleft]{threeparttable}
\usepackage{subcaption}

\definecolor{cspPurple}{RGB}{153, 0, 153}

\lstdefinelanguage{CSP}{
  morekeywords={channel,process,SKIP,STOP,WAIT,->,[],||,if,then,else,let,within,timed_priority},
  sensitive=true,
  morecomment=[l]{--},
  morestring=[b]",
}

\begin{document}
\title{A Verification Methodology for Safety Assurance of RAS}
\titlerunning{A Verification Methodology for Safety Assurance of RAS}
% If the paper title is too long for the running head, you can set
% an abbreviated paper title here
%
\author{Mustafa Adam\inst{1}\orcidID{0009-0007-8793-7435} \and
David Anisi\inst{1,2}\orcidID{0000-0003-0870-4259}
Pedro Ribeiro\inst{3}\orcidID{0000-0003-4319-4872}}
\authorrunning{M. Adam et al.}
% First names are abbreviated in the running head.
% If there are more than two authors, 'et al.' is used.
%
\institute{Faculty of Science and Technology, Norwegian University Of Life Sciences (NMBU)\\
\url{http://www.nmbu.no} \and
Department of Mechatronics, University of Agder (UiA)\\
\url{http://www.UiA.no} \and
Department of Computer Science, University Of York, UK\\
\url{https://www.york.ac.uk/} }
\maketitle              % typeset the header of the contribution
\begin{abstract}

Autonomous robots deployed in shared human environments, such as agricultural settings, require rigorous safety assurance to meet both functional reliability and regulatory compliance. These systems must operate in dynamic, unstructured environments, interact safely with humans, and respond effectively to a wide range of potential hazards. This paper presents a verification workflow for the safety assurance of an autonomous agricultural robot, covering the entire development life-cycle, from concept study and design to runtime verification. The outlined methodology begins with a systematic hazard analysis and risk assessment to identify potential risks and derive corresponding safety requirements. A formal model of the safety controller is then developed to capture its behaviour and verify that the controller satisfies the specified safety properties with respect to these requirements. The proposed approach is demonstrated on a field robot operating in an agricultural setting. The results show that the methodology can be effectively used to verify safety-critical properties and facilitate the early identification of design issues, contributing to the development of safer robots and autonomous systems.

\end{abstract}
\section{Introduction}
The deployment of Robots and Autonomous Systems (RAS) in agriculture is rapidly accelerating, offering substantial improvements in efficiency, sustainability, and productivity across diverse tasks~\cite{agricultural_robots}. As these systems transition from controlled environments to real-world agricultural settings, assuring their safety and reliability becomes essential for regulatory compliance and stakeholder trust. Dynamic and unstructured conditions—such as variable terrain, changing weather, and human presence—introduce complex safety challenges that demand robust risk mitigation and verifiable assurance frameworks~\cite{bakirtzis2023dynamic}.

Standards like ISO~18497~\cite{iso_agricultural_machinery}, IEC~61508~\cite{SafetyHandbook_smith2020} provide foundational guidance for safety-critical systems. In this context, the IEEE 7009-2022 standard~\cite{ieee7009} provides a more focused framework for addressing fail-safe design in autonomous and semi-autonomous systems. However, these standards often target conventional, deterministic systems and may not fully address the autonomy and complexity of modern robotic platforms. A key challenge, as noted by Fisher et al.~\cite{fisher2021framework}, is translating high-level, textual standards into formal, machine-verifiable specifications suitable for automated analysis.

Formal verification (FV) ensures that a system fulfils given specifications in all circumstances regardless of input possibilities~\cite{Peled2019}. FV methods extend beyond traditional testing approaches, such as unit testing or hardware-in-the-loop simulation, which lack formal guarantees and do not scale well with the increased complexity of autonomous systems~\cite{bakirtzis2023dynamic}. Luckcuck \emph{et al.} address the rationale for employing FV of RAS~\cite{Luckcuck2019}, providing a survey of the state-of-the-art on formalism and challenges. While FV methods like Model Checking (MC) and theorem proving offer rigorous verification, many existing approaches focus solely on offline or static analysis, limiting their applicability in real-world scenarios. In fact, a very recent, structured literature review regarding use of FV methods for RAS, which has also been submitted to TAROS~\cite{AzaiezSurvey2025}, reveals that merely 8\% of all relevant papers consider an "integrated approach", combining offline MC and online Runtime Verification (RV).

This paper presents a formal verification (FV) methodology for RAS that combines formal modelling, safety requirement traceability, and explicit treatment of uncertainty across the engineering lifecycle. Our approach supports the verification of safety properties from hazard analysis through to runtime verification and operational assessment. To complete an important but missing piece in our earlier works~\cite{adam2023case,Adam2024CASE},  we place particular emphasis on the Safety Controller (SC) in this paper, a key component responsible for enforcing runtime safety responses.

The remainder of this paper is structured as follows: Section~\ref{sec:methodology} introduces the overall verification methodology. Section~\ref{sec:sis} details the formal modelling and verification of the Safety Controller. Finally, Section~\ref{sec:conclusion} concludes the paper and highlights directions for future work.

%\section{Overview of the Framework}
%\label{sec:framework}
\section{Formal Verification Methodology}
\label{sec:methodology}
This section introduces the overall verification methodology and outlines the activities for the safety assurance of RAS. An agricultural robot performing UVC-light treatment in strawberry plants is used as a representative case study to illustrate the application of the methodology~\cite{adam2023case}. The proposed approach builds upon the workflow introduced in~\cite{Mario_2020}. While the specific hazards addressed in the case study pertain to UVC treatment, the methodology itself is generalizable and applicable to a broad range of RAS.

The figure below illustrates the lifecycle-oriented verification methodology, detailing each engineering phase and its verification-specific steps within the broader development process. It also highlights complementary engineering activities to show how verification is integrated throughout the system lifecycle. 
\newline 
Next, we detail the workflow, activities and artifacts of each engineering phase.
\begin{figure}[htbp!]
    \centering
    \includegraphics[width=\linewidth]{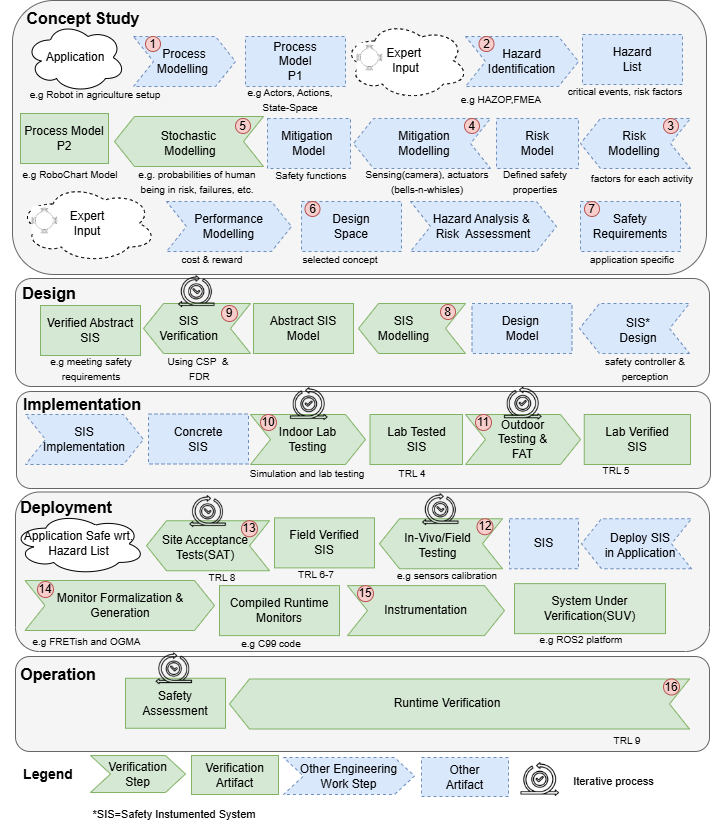}
    \caption{Overview of the verification methodology across the engineering lifecycle. Iterative steps allow returning to the relevant previous phase based on the failure’s root cause. All artifacts shown have been produced and exemplified in our agricultural case study.}
    \label{fig:methodology}
\end{figure}
% No need for a new paragraph?
 
\subsection{Concept Study}
The concept-study phase, which for larger projects may occur several years before system deployment and operation, lays the foundation for safety verification by defining the application, identifying hazards, modelling processes, and evaluating risks. It combines expert domain knowledge with formal analysis to extract critical safety requirements. 
\subsubsection{Process Modelling} \Circled[inner color=blue]{1} The workflow commences by modelling a high-level model of the robotic application, capturing its interaction with the environment, operational constraints, and domain-specific requirements. In our case study, the scenario involves an autonomous robot performing UVC treatment in strawberry crops — a task that underscores the safety nature of the system, as exposure of the human eye to the UVC light may lead to injuries. The model incorporates key \textbf{actors} (e.g., robot, human - trained or untrained), core \textbf{activities} (such as a human approaching a robot), \textbf{actions}, and UVC light as the main source of risk. Taken together,  this results in the definition of the \textit{Process Model}.

\subsubsection{Hazard Identification}  \Circled[inner color=blue]{2} This step involves a systematic exploration of potential sources of harm arising from the robot’s behavior or its interaction with the environment. Common techniques adopted include Hazard Identification (HAZID) and Hazard and Operability Study (HAZOP), which combines domain-specific models with structured hazard analysis to inform early safety decisions,  Failure Mode and Effects Analysis (FMEA), and What-If Analysis~\cite{SafetyHandbook_smith2020}. These structured engineering approaches help uncover failure modes, operational risks, and unsafe conditions early in the development process.

In our case study, we adopt the HAZOP approach informed by domain expertise and a structured evaluation of operational scenarios~\cite{adam2023case,guevara2021uvc}. The process begins with the identification of potential failure modes and their associated consequences, followed by a qualitative risk assessment using a standard risk matrix. The process is driven by expert input, and it typically follows a structured but collaborative process involving domain knowledge and safety engineering practices.

The output is a curated \textit{Hazard List} detailing unsafe conditions relevant to the UVC application.

\subsubsection{Risk Modelling}
 \Circled[inner color=blue]{3} After identifying hazards, risks are assessed based on likelihood and severity to prioritize safety requirements. Approaches such as Preliminary Risk Analysis (PRA), probabilistic risk modelling, and activity-based risk graphs can be applied. In our work, we utilize activity-based risk modeling with a custom matrix that captures the likelihood of human presence, sensor failure rates, and robot behavior patterns. This results in a Risk Model that supports formal requirement derivation and highlights a critical scenario: the robot, during UVC treatment, fails to detect a human entering a hazardous zone during row transition.

\subsubsection{Mitigation Modelling}
\label{sec:methodology_mitigation}
\Circled[inner color=blue]{4} Based on identified hazards and associated risk-level, \textit{Mitigation Models} specify system reactions to risks. These are used to shape a design space where alternative mitigation actions are analysed to achieve the desired Safety Integrity Level (SIL). The output is a detailed mitigation model which will be integrated later in the safety controller design.

\subsubsection{Stochastic Modelling} \Circled[inner color=blue]{5} Probabilistic aspects such as human intrusion, sensor failure, and actuator uncertainty are captured in a \textit{Process Model}, developed using RoboChart’s~\cite{robochart} probabilistic modelling facilities. The robot’s UVC treatment behaviour, the object detection system (ODS), and human presence are each represented as individual probabilistic state machines synchronised through a discrete-time tick mechanism~\cite{adam2023case}. The combined model is automatically translated into PRISM's input language. Formal properties, such as the probability of a human being exposed to UVC light in a danger zone during the transition, are verified using PCTL queries in PRISM. In \Circled[inner color=blue]{6} we provide estimates for cost and reward of the different mitigation approaches, their disruption of the UVC treatment process, and the effort required for their execution.

\subsubsection{Safety Requirements} \Circled[inner color=blue]{7} Here we formalise the system level requirements based on the SIL quantification from previous steps. The level of risk ranging from SIL 1 to SIL 4, each increasing level corresponds to a 10-fold decrease in Probability of Failure on Demand (PFD), hence the higher the SIL, then the more stringent requirements. Our probabilistic model checking enables estimation of whether system configurations (e.g., sensor performance or human awareness levels) can meet target SIL thresholds.
 
\subsection{Design Phase}
In the design phase of our methodology, safety requirements are translated into concrete component-level specifications based on identified hazards and risk analyses. The Safety Instrumented Functions (SIFs) is designed, incorporating sensors, internal logic, and actuators to fulfil  the  required SILs. In \Circled[inner color=blue]{8} we model the SIS using RoboChart to represent the system as independent processes that interact via message-passing communication channels. In step \Circled[inner color=blue]{9}, offline model-checking is used to verify formally these designs with respect to the defined safety requirement before implementation begins, ensuring early-stage correctness.

\subsection{Implementation Phase}

The verified design is implemented as integrated hardware and software components on a Thorvald platform robot. The SIS is realized using ROS2 for the controller, and the Obstacle Detection System (ODS) is based on a YOLO model running with input from an Intel RealSense D435 camera. These choices are direct consequences of the verification and analysis results obtained during the earlier concept-study and design phases. Their performance, however, needs to be tested and verified in increasingly more realistic settings.

In \Circled[inner color=blue]{10}, we conduct indoor lab testing to empirically assess the classification model's accuracy. The dataset is partitioned into training, testing, and validation sets, achieving a classification accuracy of 94\%, well above the 70\% threshold defined in the concept study to achieve the desired SIL~\cite{adam2023case}. Additionally, the SIS is simulated using ROS2 and Gazebo, where Gazebo serves as the virtual environment for the robot. Custom plugins were developed to enable bi-directional interaction between the Gazebo simulation and the ROS2 nodes. The goal of this stage is to achieve Technology Readiness Level (TRL) 4.

In \Circled[inner color=blue]{11}, outdoor testing is performed in the intended deployment environment. This phase confirms the system’s performance. The outcome of both indoor and outdoor lab testing is a SIS, demonstrating compliance with the system requirements and achieving TRL 5.

\subsection{Deployment Phase}

\begin{figure}[hbtp!]
    \centering
    \includegraphics[width=0.45\textwidth]{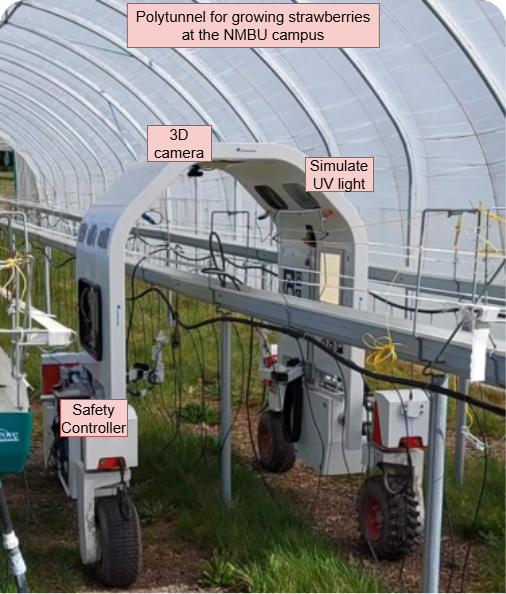}
    \caption{The robot was deployed in a polytunnel for simulated UVC operation.}
    \label{fig:thorvaldNMBU}
\end{figure}

The SC is deployed on a real agricultural robot and evaluated in dynamic outdoor environments to ensure its performance under real-world conditions, including varying lighting, terrain, and environmental dynamics. The setup is presented in Figure~\ref{fig:thorvaldNMBU}.

In \Circled[inner color=blue]{12}, the sensors are calibrated according to the mission profile, and the robot is deployed for prototype demonstration in relevant environments. Testing is conducted in intended operational conditions, ideally at the final deployment site. The focus at this stage is on ensuring correct integration, and confirming basic functionality — such as in open fields that mimic actual farms. This on-site verification of SIS raises the TRL to 6-7.

In \Circled[inner color=blue]{13}, the complete system is tested and accepted at the actual operational site. This stage emphasizes the assessment of the system's reliability, robustness, and compliance with operational and safety requirements under real deployment conditions. The system tested here is the production-ready version, with all components finalized and integrated for use in the real world. Conducting Site Acceptance Test (SAT) yields TRL of 8.
In \Circled[inner color=blue]{14} we generate runtime monitors from formalized properties of interest. We used the Formal Requirement Elicitation Tool (FRET)~\cite{FRET}, to capture requirements in structured natural language and automatically translate into monitor specifications. These are then processed by OGMA~\cite{NASA_OGMA_2024} to generate C99 monitor code, which can be synthesized and integrated in ROS 2 platform in step \Circled[inner color=blue]{15}.

\subsection{Operation Phase}
In the operation phase, the robot performs its agricultural tasks in a real-world environment. To ensure compliance with safety requirements at runtime, a runtime verification (RV) framework is utilized~\cite{Adam2024CASE}. It employs monitors generated from formal safety specifications to observe system behavior and verify adherence to safety properties. In \Circled[inner color=blue]{14}, the RV framework evaluates both temporal and state-based safety conditions during live operation. Upon detecting potential or actual violations, it triggers predefined mitigation actions, such as halting motion, issuing alerts, or transitioning the system into a safe state. This proactive mechanism enhances operational safety, particularly in dynamic and uncertain field environments. Step \Circled[inner color=blue]{15} provides an assessment and feedback loop leveraging runtime data to uncover unexpected scenarios, edge cases, and near-misses. This feedback is used both to evaluate the effectiveness of the proposed methodology and iteratively improve the system’s models, safety requirements, and implementation, thus supporting a continuous and adaptive safety assurance lifecycle.

\section{Safety Controller (SC)}
\label{sec:sis}
A \textit{Safety Controller} ensures system safety by monitoring conditions and executing mitigation actions when necessary. Unlike general-purpose control systems, it acts independently to detect hazards and prevent unsafe behavior, even with component faults or degraded performance~\cite{iec61508}.

In autonomous agricultural robotics, SC is vital due to the complex and dynamic nature of farm environments. Terrain, weather, lighting, unexpected human or animal presence, and hazardous equipment demand context-aware safety responses. The controller evaluates situational data to manage risks in real-time. However, agricultural robots must also maintain high operational efficiency to perform time-sensitive, energy-intensive tasks. Overly conservative safety actions, like unnecessary stops, can lower throughput, waste energy, and erode trust. The core challenge is balancing safety with efficiency—designing controllers that respond proportionally to context and risk is key to scalable deployment
\subsection{Modelling in RoboChart}
\begin{figure}[hbtp!]
    \centering
    \includegraphics[width=\textwidth]{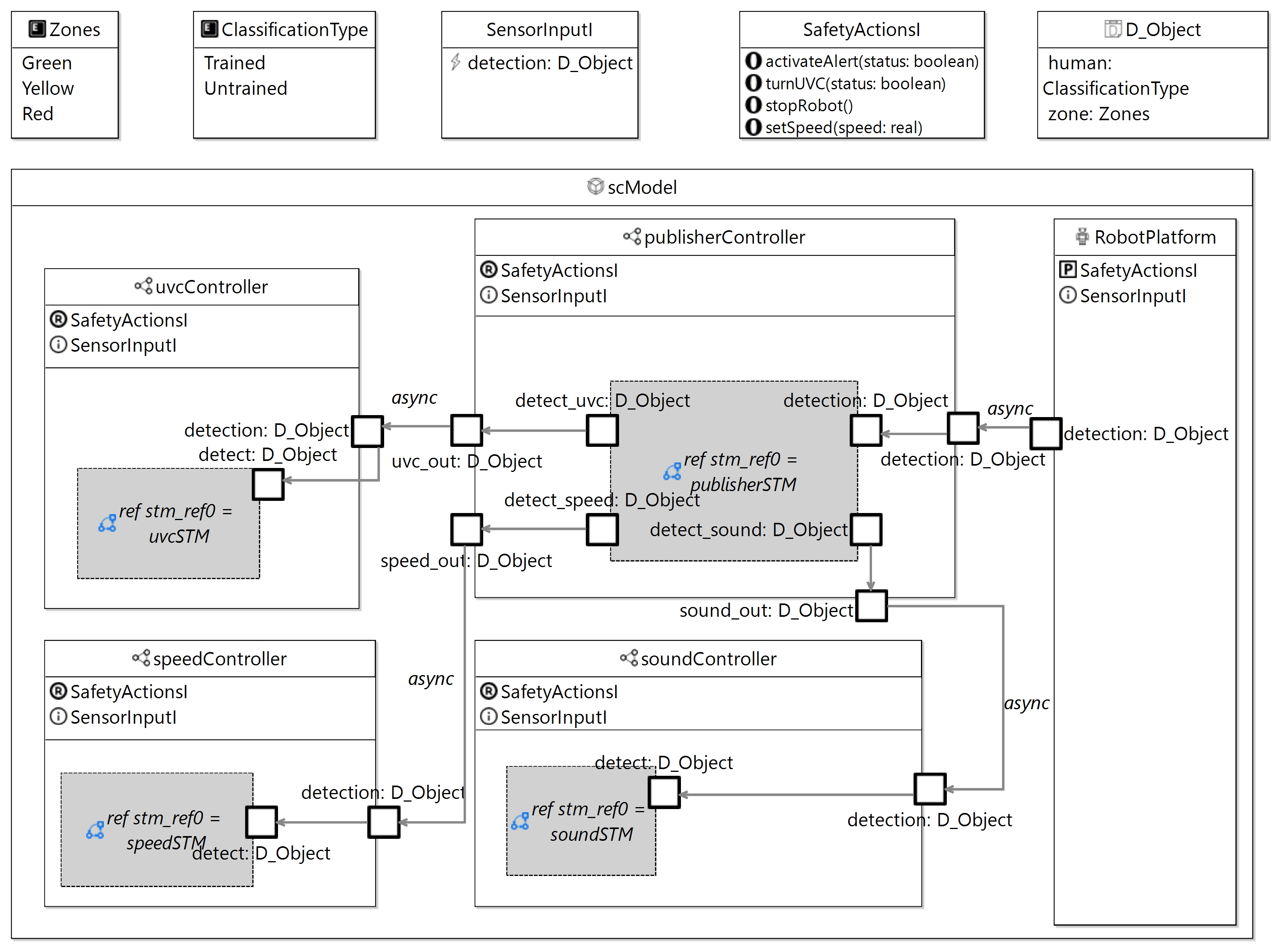}
    \caption{Overview of the RoboChart model of the SIS architecture, generated using RoboTool.}
    \label{fig:robocchart-model}
\end{figure}

This section details the formal modeling and verification of the SIS safety controller using RoboChart~\cite{robochart}, a state-machine-based formalism for reactive robotic systems. Abstraction and modularization techniques manage complexity and ensure all critical safety requirements are covered. The model is verified through refinement checks to confirm compliance with safety properties before deployment. The RoboChart model in Figure~\ref{fig:robocchart-model} reflects the system architecture while abstracting hardware details~\cite{MURRAY2022102766}.

The robotic platform, denoted as \texttt{RobotPlatform}, defines the external interfaces available to the safety software. It uses the \texttt{SensorInputI} interface to receive environment-derived classification events from perception modules, and provides the \texttt{SafetyActionsI} interface for invoking critical mitigation actions, such as \texttt{activateAlert}, \texttt{turnUVC}, \texttt{stopRobot}, and \texttt{setSpeed}. These operations originate from the mitigation strategies introduced in Section~\ref{sec:methodology_mitigation}.

To support semantic classification of detected objects, we define two enumerated types: \texttt{Zones} (with values \texttt{Green}, \texttt{Yellow}, \texttt{Red}) and \texttt{ClassificationType} (with values \texttt{Trained}, \texttt{Untrained}). These are composed into a record type \texttt{D\_Object}, representing classified detections from the perception system and used as inputs to the RoboChart controllers.

\begin{figure}[hbtp!]
    \centering
    \includegraphics[width=\textwidth]{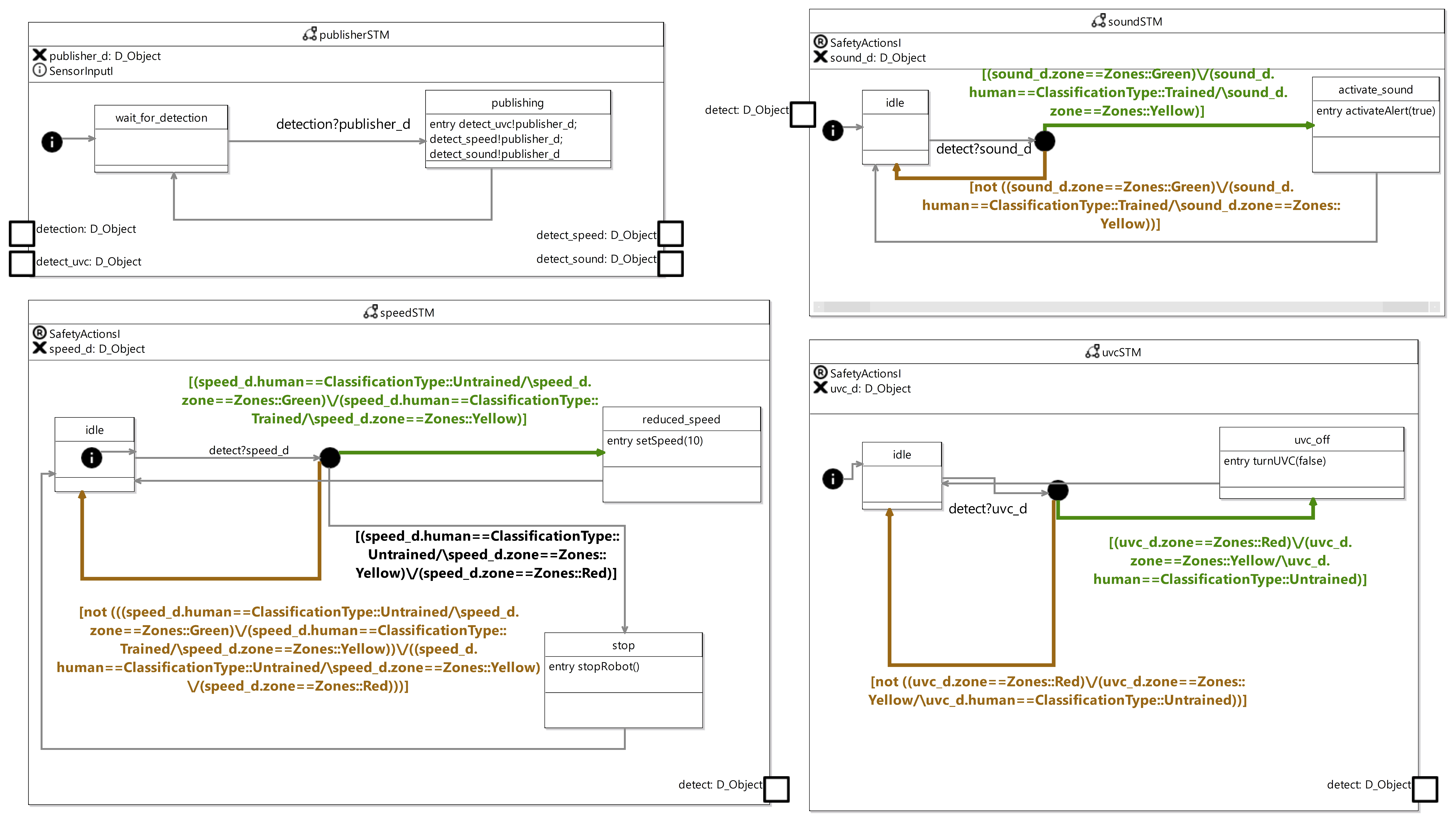}
    \caption{State machines defining the safety response logic for UVC control, speed management, and alerting (sound).}
    \label{fig:robocchart-statemachines}
\end{figure}

Figure~\ref{fig:robocchart-model} is composed of four interacting controllers: \texttt{publisherController}, \texttt{uvcController}, \texttt{speedController}, and \texttt{soundController}. The publisherController serves as a dispatcher that receives detection events and asynchronously relays them to the remaining controllers via typed events. Each controller encapsulates a state machine that governs its decision-making logic. The state machines are shown in Figure~\ref{fig:robocchart-statemachines}. The \texttt{uvcSTM} disables the UVC light when a human is detected in high-risk proximity zones, depending on their classification. The \texttt{speedSTM} either reduces the robot’s velocity or halts it based on proximity and human training level. The \texttt{soundSTM} triggers alerts in lower-risk situations to increase awareness. This illustrates the balancing act between risk level and safety on one hand, and operational efficiency on the other.

Transitions between states are triggered by events and guarded by Boolean conditions over the fields of \texttt{D\_Object}. Logical conditions are derived from hazard mitigation rules and encoded into transition guards, ensuring deterministic and safe behaviour. This structure follows the separation-of-concerns principle advocated in~\cite{murray2024model}, where physical execution is abstracted into service operations and software safety logic is isolated in controller-specific state machines. The verification process is discussed in Section~\ref{sec:verification}.

%%All interactions in the model are explicitly defined using RoboChart’s asynchronous event channels and interface contracts. This modular and layered architecture facilitates automatic generation of CSP-based semantic models, enabling the use of RoboTool and FDR for formal verification of safety properties and deadlock-freedom~\cite{robochart, robostar}.

%%This modelling with the assertion ensures traceability to safety requirements and serves as a verifiable foundation for runtime enforcement of mitigation actions. 

\subsection{Verification of Safety Properties}
\label{sec:verification}

To ascertain that the RoboChart model complies with the safety requirements, we first capture the requirements as timed behaviours in $tock$-CSP~\cite{BaxterRC22}, a discrete timed process algebra. Then, we verify that the behaviour of the RoboChart model, whose formal semantics is defined as a $tock$-CSP process, is that permitted by the specification via the notion of refinement that captures conformance. For automated verification, we leverage RoboTool's integration with the CSP~\cite{Roscoe10} model-checker FDR~\cite{FDR}, following the methodology outlined in~\cite{robochart}.

%specify each requirement as a $tock$-CSP~\cite{} process and use refinement to check that the model satisfies the desired behaviors, namely via RoboTool's integration with the CSP model-checker FDR~\cite{}. Following the methodology outlined in~\cite{robochart}, our approach consists of specifying expected safety responses as abstract timed reactive process models and then checking whether the $tock$-CSP semantics automatically calculated for a RoboChart model by RoboTool refines the process model.

We focus on verifying Safety Requirements, \textbf{R1} to \textbf{R5} as specified in Table~\ref{table:hazards_cases}, which require that an appropriate mitigation action must occur within a bounded time interval upon the detection of a human in the robot’s vicinity. Each case specifies one or more actions (e.g., alert, speed reduction, stopping, or UVC shutdown) that must be completed within a defined deadline. Next, we illustrate how the requirements are captured in $tock$-CSP.
\begin{table}[hbtp!]
  \centering
  \caption{Safety Requirements, $R_i$, and mitigation actions.}
  \begin{threeparttable}
  \begin{tabular}{|p{0.4cm}|p{3cm}|p{1cm}|p{4cm}|}  
    \hline
   \textbf{ID} & \textbf{Human detected} & \textbf{Zone} & \textbf{Mitigation Actions} \\
 % \hline    
 %  R0 & No human is detected 
 %   & Nothing \\
   \hline
   $R_1$ & Trained  
   & Green 
    & Activate sound  \\
  
    $R_2$ & Untrained 
    &Green
    &Activate sound \& slow down \\

    $R_3$ & Trained 
    &Yellow 
    & Activate sound \&  slow down  \\
    $R_4$ & Untrained
    &Yellow
    & Turn off UVC \& Stop robot\\
        
    $R_5$& Trained/Untrained
    &Red
    & Turn off UVC \& Stop robot  \\
    \hline
  \end{tabular}   
  \end{threeparttable}
 \label{table:hazards_cases}  
\end{table}
Listing~\ref{lst:spec-r1} sketches the process \lstinline{SpecR15}, which captures the safety requirements using CSP events by describing the required pattern of interaction via a set of equational definitions. Here, we capture all requirements in a single process, so that they can all be checked at once, although they could have equivalently been captured via separate processes. As the requirements are timed, \texttt{SpecR15} is defined within a \lstinline{Timed} section \lstinline|{...}| and uses a function \lstinline{timed_priority} to ensure the correct timed semantics are calculated by FDR\footnote{\href{https://cocotec.io/fdr/manual/cspm/definitions.html}{https://cocotec.io/fdr/manual/cspm/definitions.html}}. Initially, it behaves as specified by \lstinline{Initial}, that accepts an input (\lstinline{?}) event \lstinline{scModel::detection.in} with an arbitrary value \lstinline{i} typed according to the RoboChart event \lstinline{detection}, and then (\lstinline{->}) behaves as \lstinline{Response(i,2)} followed (\lstinline{;}) by the recursion, that handles a new input. Here, \lstinline{i} is the value communicated and 2 the maximum number of time units for the response to occur. The \lstinline{Response} is defined by another process \lstinline{within} a context where \lstinline{h} and \lstinline{z} take the value of the human and zone components of \lstinline{i} in a \lstinline{let} definition. Then, there is an \lstinline{if} case for each possible input. In Listing~\ref{lst:spec-r1} we sketch the first two requirements and omit the others as they are of a similar nature. For \textbf{R1}, a \lstinline{Green} zone detection of a \lstinline{Trained} person triggers an alert, here specified using a deadline operator (\lstinline{EndBy}) that requires that the process in the first argument, representing the call to the operation \lstinline{activateAlert} with value \lstinline{True}, terminates (\lstinline{SKIP}) within \lstinline{d} time units. The construction is similar for \textbf{R2}, but instead the response triggers both \lstinline{activateAlert} and \lstinline{setSpeed}, so in that case we use interleaving (\lstinline{|||}) as their order is irrelevant. % TBC
%
%using the \texttt{EndBy} operator to specify that each required response must occur within $d=2$ time units after detection. 
%For instance, a \texttt{Green} zone detection of an \texttt{Untrained} person triggers both an alert and a speed reduction, encoded as parallel deadline-constrained events.
%
\begin{lstlisting}[language=CSP, caption={Sketch of $tock$-CSP specification of safety requirements R1 and R2.}, label={lst:spec-r1}]
timed csp SpecR15 csp-begin
Timed(OneStep) { -- For calculation of tock-CSP semantics
 SpecR15 = timed_priority(Initial)
 -- Top-level specification process
 Initial = scModel::detection.in?i -> Response(i,2) ; Initial
 -- Response after an input detection is received
 Response(i,d) = (
  let h = D_Object_human(i)
      z = D_Object_zone(i)
  within -- One case per input value
  if (z==Zones_Green and h==ClassificationType_Trained) 
   then (EndBy(scModel::activateAlertCall.True -> SKIP,d))
  else if(z==Zones_Green and h==ClassificationType_Untrained) 
   then (EndBy(scModel::activateAlertCall.True -> SKIP,d) 
         ||| EndBy(scModel::setSpeedCall.10 -> SKIP,d))
    ... ) } csp-end
\end{lstlisting}
For verification, we consider the RoboChart model in a context where we assume that the \lstinline{detection} input does not change more often than once every time unit to reflect realistic sensing intervals. This is captured in $tock$-CSP by the  process \lstinline{CSc}, defined in~Listing~\ref{lst:constrained-model}, as the parallel composition (\lstinline{[|...|]}) of \lstinline{scModel::O__}, the $tock$-CSP semantics automatically calculated by RoboTool for the RoboChart model, and \lstinline{Inp} synchronising on the channel set (\texttt{\{|...|\}}) that includes the event \texttt{detection} with any value. In process \lstinline{Inp}, we require that at least one time unit elapses (\lstinline{WAIT(1)}) between each \lstinline{detection} input.

%The concrete system model generated from RoboChart is defined in the \texttt{CSc} process (Listing~\ref{lst:constrained-model}), which synchronises the \texttt{scModel} with a detection input sequence and introduces a delay of 1 time unit between inputs to reflect realistic sensing intervals.

\begin{lstlisting}[language=CSP, caption={System model with constrained timed input for refinement checking.}, label={lst:constrained-model}]
timed csp CSc associated to scModel csp-begin
Timed(OneStep) {
 CSc = timed_priority(
  let Inp = scModel::detection.in?x__ -> WAIT(1) ; Inp
      Def = scModel::O__(1) [|{|scModel::detection.in|}|] Inp
  within Def) }
csp-end
\end{lstlisting}
We then formulate a refinement \lstinline{assertion} \lstinline{A1}, as shown in Listing~\ref{lst:assertion}, to check whether the system behavior of \lstinline{CSc} \lstinline{refines} the expected specification \lstinline{SpecR15} in the CSP \lstinline{traces} semantic model, that is adequate for reasoning about safety and timeliness. Here, safety means that all possible timed executions of the implementation are allowed by the safety specification (\lstinline{SpecR15}). 
\begin{lstlisting}[language=CSP, caption={Timed refinement assertion for safety verification.}, label={lst:assertion}]
timed assertion A1 : CSc refines SpecR15 in the traces model.
\end{lstlisting}
In addition to this refinement check, we verify key correctness properties of the RoboChart model, including \textit{deadlock freedom} and \textit{determinism}. Such assertions can be written effortlessly without having to define a CSP process, and similarly to the previous assertion are checked using RoboTool's integration with FDR. 
%All assertions are automatically checked using the FDR model-checker.

In our particular example, due to the relatively small size and complexity, the model was successfully verified within a few seconds with all assertions passing and, therefore, no counterexamples generated. This confirms that safety mitigation actions are correctly triggered and completed within the defined 2-unit time bound after a corresponding detection event. The verification was completed within a few seconds and involved a modest number of states and transitions, posing no scalability issues. These properties ensure that the system does not enter a state from which no further progress is possible and that its behaviour is deterministic, that is, a particular detection input leads to a specific response.%, thus remaining predictable. % remains predictable and unambiguous in the presence of concurrent events.

\section{Conclusion}
\label{sec:conclusion}

This work presents an integrated, methodology for RAS verification that combines probabilistic modelling, formal methods, and RV across the engineering lifecycle. Using RoboChart, we formally model and verify the SIS, with safety properties encoded in $tock$-CSP and verified via FDR through RoboTool. Evaluating the methodology's effectiveness involves both adherence to established standards~\cite{SafetyHandbook_smith2020} and empirical perfzormance in an actual operation setting. While alignment with best practices and workflows builds confidence, real-world performance—captured through RV, near-misses, and safety incidents—ultimately determines success. In our methodology, these steps have been incorporated in the Operation phase (see \Circled[inner color=blue]{15}), where operational data is used to evaluate the effectiveness of the overall methodology and iteratively refine the models, further strengthening assurance over time.
%Throughout, a specific use case from agriculture robotics has served as a motivating and illustrative example. The outlined methodology however, can be applied to other applications and domains involving autonomous and safety-critical systems where timely response and verifiable behaviour are essential. Its strength lies in the integration of static design-time guarantees with dynamic operational monitoring, closing the loop between formal modelling and practical deployment.
This step also helps addressing a challenge within all model-based methods, namely  ensuring that the formal model accurately reflects the real-world system. While abstractions are needed to manage complexity, they must be carefully balanced to avoid oversimplification and minimise the \textit{reality gap} between the model and the deployed system.

Future work will focus on advancing from online RV to \textit{predictive} RV, where monitors proactively anticipate potential hazards based on system and environmental state~\cite{zhang2012runtime,pinisetty2017predictive}. This predictive power enables earlier mitigation and deeper integration with autonomous decision-making, enhancing overall system safety and responsiveness.

\begin{credits}
\subsubsection{\ackname} The research presented in this paper has received partial funding from the Norwegian Research Council (RCN) RoboFarmer, project number 336712, the UK EPSRC Grants EP/M025756/1, EP/R025479/1, and EP/V026801/2, and  the Royal Academy of Engineering Grant No CiET1718/45.

\subsubsection{\discintname}
The authors have no competing interests to declare
that are relevant to the content of this article..
\end{credits}
%
% ---- Bibliography ----
%
% BibTeX users should specify bibliography style 'splncs04'.
% References will then be sorted and formatted in the correct style.
%
% \bibliographystyle{splncs04}
% \bibliography{mybibliography}
%

\bibliographystyle{splncs04}
\bibliography{bibliography}

\end{document}